\documentclass[runningheads]{llncs}
\usepackage{graphicx}
\usepackage{tikz}
\usepackage{comment}
\usepackage{amsmath,amssymb} 

\usepackage[accsupp]{axessibility}  

\usepackage{booktabs}

\usepackage[pagebackref,breaklinks,colorlinks]{hyperref}

\usepackage{amsfonts}
\usepackage{dsfont}
\usepackage{threeparttable}
\usepackage{color,colortbl}
\usepackage{pifont}
\newcommand{\cmark}{\ding{51}}%
\newcommand{\xmark}{\ding{55}}%
\usepackage{multirow}
\usepackage{multicol}
\usepackage{diagbox}
\usepackage{arydshln}
\usepackage{algorithm}
\usepackage{float}
\usepackage{wrapfig}
\usepackage[misc]{ifsym}
\usepackage{soul}

\definecolor{Gray}{gray}{0.85}
\newcolumntype{a}{>{\columncolor{Gray}}c}
\definecolor{blush}{rgb}{0.87, 0.36, 0.51}

\usepackage[capitalize]{cleveref}
\crefname{section}{Sec.}{Secs.}
\Crefname{section}{Section}{Sections}
\Crefname{table}{Table}{Tables}
\crefname{table}{Tab.}{Tabs.}

\begin{document}

\pagestyle{headings}
\mainmatter
\def\ECCVSubNumber{5922}  

\title{BMD: A General Class-balanced Multicentric Dynamic Prototype Strategy for Source-free Domain Adaptation} %

\titlerunning{BMD: General Strategy for Source-free Domain Adaptation}
%
\author{Sanqing Qu\inst{1} \and Guang Chen\inst{1}$^{(\textrm{\Letter})}$ \and Jing Zhang\inst{2} \and Zhijun Li \inst{3}\and \\
Wei He\inst{4} \and Dacheng Tao\inst{2,5}}
\authorrunning{Sanqing et al.}
%
\institute{Tongji University \email{\{2011444, guangchen\}@tongji.edu.cn} \and The University of Sydney  \email{jing.zhang1@sydney.edu.au}
\and University of Science and Technology of China \email{zjli@ieee.org} \and University of Science and Technology Beijing \email{weihe@ieee.org}  \and JD Explore Academy 
\email{dacheng.tao@gmail.com}}
\maketitle
\begin{abstract}
Source-free Domain Adaptation (SFDA) aims to adapt a pre-trained source model to the unlabeled target domain without accessing the well-labeled source data, which is a much more practical setting due to the data privacy, security, and transmission issues. To make up for the absence of source data, most existing methods introduced feature prototype based pseudo-labeling strategies to realize self-training model adaptation. However, feature prototypes are obtained by instance-level predictions based feature clustering, which is category-biased and tends to result in noisy labels since the visual domain gaps between source and target are usually different between categories. In addition, we found that a monocentric feature prototype may be ineffective to represent each category and introduce negative transfer, especially for those hard-transfer data. To address these issues, we propose a general class-\textbf{B}alanced \textbf{M}ulticentric \textbf{D}ynamic prototype (BMD) strategy for the SFDA task. Specifically, for each target category, we first introduce a global inter-class balanced sampling strategy to aggregate potential representative target samples. Then, we design an intra-class multicentric clustering strategy to achieve more robust and representative prototypes generation. In contrast to existing strategies that update the pseudo label at a fixed training period, we further introduce a dynamic pseudo labeling strategy to incorporate network update information during model adaptation. Extensive experiments show that the proposed model-agnostic BMD strategy significantly improves representative SFDA methods to yield new state-of-the-art results. The code is available at \url{https://github.com/ispc-lab/BMD}.
\end{abstract}

\section{Introduction}
\label{sec:intro}

\begin{figure}[ht]
    \centering
    \includegraphics[width=0.70\textwidth]{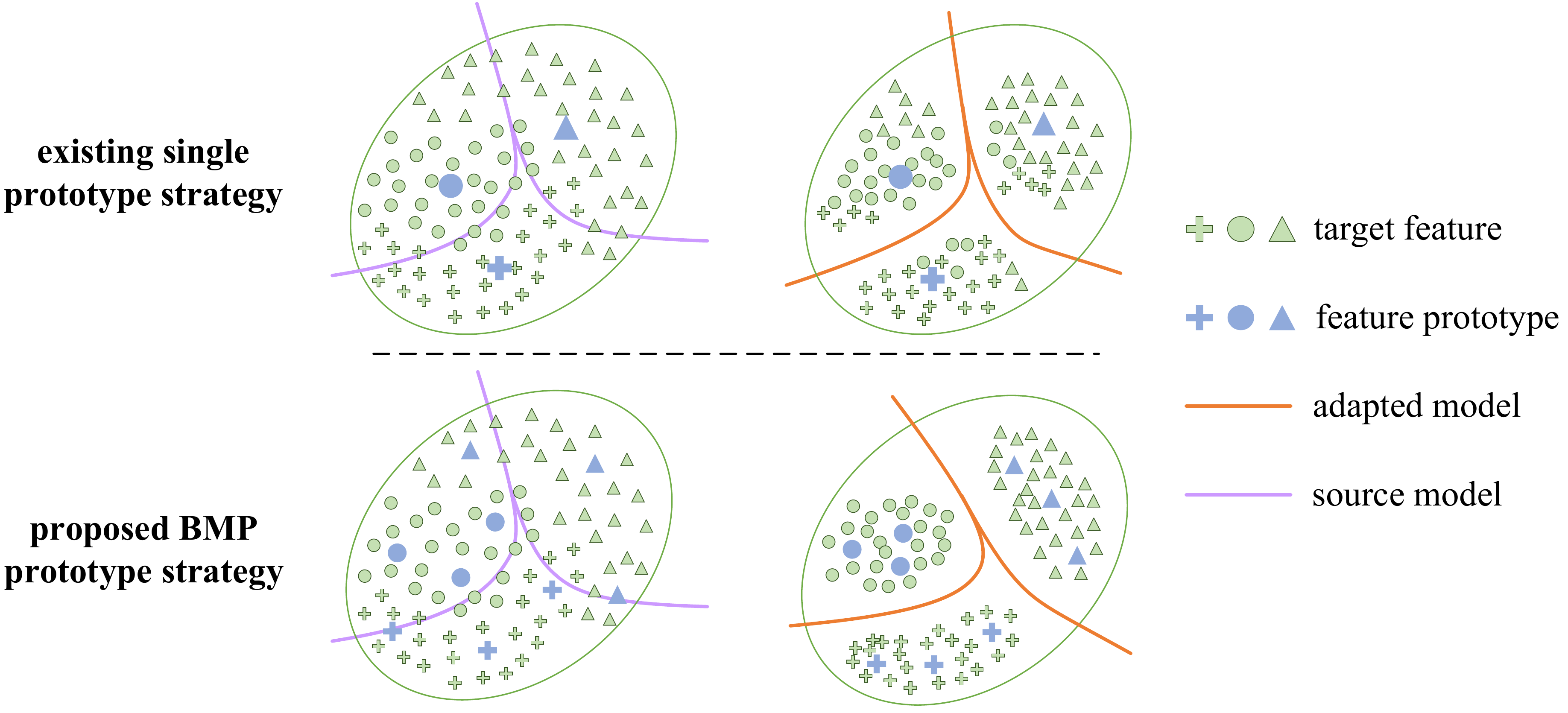}
    \caption{Comparison between existing prototype strategy (left) and our BMD prototype strategy (right). During SFDA model adaptation, existing prototype strategies are monocentric and class-biased, which often lead to negative transfers for those hard-transfer instances, while our class-balanced multicentric dynamic prototype (BMD) strategy can effectively address the issue.}
    \label{fig:method_comparison}
\end{figure}

\par Deep neural networks have achieved remarkable success in various visual tasks at the expense of massive data collections and annotations~\cite{imagenet,resnet,dosovitskiy2020image,xu2021vitae,ispc:vcanet21} but still often generalized poorly to the unseen new domains due to the inter-domain discrepancy. To reduce the annotation burden when dealing with new domain data, unsupervised domain adaptation (UDA) methods have been developed by aligning the well-labeled source data and the unlabeled target data distribution, which have achieved promising results in object recognition~\cite{ADDA,dann,MCD,mdd}, object detection~\cite{da_object_detection_1,wang2021exploring,da_object_detection_review}, and semantic segmentation~\cite{cbst,crst,zhang2019category,proda,gao2021dsp}. 

\par However, most existing UDA methods require to access the source and target data simultaneously during model adaptation, which is often impractical due to the concerns about data privacy, data security and data transmission efficiency. Therefore, current frontiers have emerged a few works~\cite{3c-gan,gan_sfda,gan_sfss,shot,shot++,gsfda,nrc,ms_sfda} seeking to realize source-free domain adaptation (SFDA), where only a source pretrained model is available. To make up for the absence of source data, existing methods can be divided into two main categories: GAN based methods~\cite{3c-gan,gan_sfda,gan_sfss} and self-training based methods~\cite{shot,shot++,gsfda,nrc,ms_sfda}. For those self-training methods, pseudo-labeling strategies based on feature prototypes are popular and offer promising results.~\cite{shot,shot++,ms_sfda} introduce a weighted k-means clustering based feature prototype generation strategy. However, existing strategies are implemented with instance-level prediction results, which are category-biased and tend to introducing noisy labels, since the visual (e.g. scale, appearance, etc) domain gaps between source and target are usually different between categories~\cite{cbst,crst}. In addition, we argue that due to the domain gap, a rough monocentric feature prototype for each category could not effectively represent the target data and would introduce negative transfer, especially for those hard-transfer data.

\par In light of the above issues, in this paper, we focus on existing self-training based SFDA methods and propose a general class-\textbf{B}alanced \textbf{M}ulticentric \textbf{D}ynamic (BMD) prototype strategy. Specifically, to avoid the gradual dominance of easy-transfer classes on prototype generation, for each target category we first introduce a novel inter-class balanced sampling strategy to aggregate the potential and representative data samples. Even though we can obtain category balanced feature prototype with above strategy, it is still inferior for those hard transfer data samples. Therefore, we then introduce an intra-class multicentric clustering strategy to generate multiple feature prototypes for each category to assign more robust and precise pseudo labels. In addition, we conjecture that existing strategies that update the pseudo label bank at a fixed training period, may not effectively exploit the dynamic information of network optimization. Thus, we further introduce a dynamic pseudo-labeling strategy to incorporate network update information during model adaptation. We compare our BMD strategy with existing methods in Fig.~\ref{fig:method_comparison}. 

\par To evaluate the effectiveness and generality of our model agnostic strategy, we have applied our strategy to four existing representative methods (SHOT~\cite{shot}, SHOT++~\cite{shot++}, G-SFDA~\cite{gsfda} and NRC~\cite{nrc}). Extensive experiments on four benchmark datasets (VisDA-C~\cite{visda}, Office-Home~\cite{officehome}, Office-31~\cite{office31} and PointDA-10~\cite{pointda}) show that our BMD strategy significantly improves these methods to yield new state-of-the-art performance.

\par Our contribution can be summarized as follows:
\begin{itemize}
    \item We propose a general class-balanced multicentric dynamic prototype strategy BMD for SFDA tasks that is model-agnostic and can be applied to existing self-training based SFDA methods.
    
    \item To avoid the gradual dominance of easy-transfer classes on prototype generation, we propose a novel inter-class balanced sampling strategy to aggregate potential and representative data samples.
    
    \item To reduce the noisy labels for those hard-transfer data samples, we introduce an intra-class multicentric prototype strategy for each category to assign more robust and precise pseudo labels.
    
    \item We conducted extensive experiments to evaluate the effectiveness of our BMD strategy. The results show that the proposed strategy can significantly boost existing methods, e.g., improving SHOT~\cite{shot} from 82.9\% to 85.8\% on VisDA-C and NRC~\cite{nrc} from 52.6\% to 57.0\% on PointDA-10.
\end{itemize}

\section{Related Work}
\label{sec:related}
\subsection{Unsupervised Domain Adaptation}
\par In pursuit of transfering knowledge from a different but well-labeled source dataset to an unlabeled but relevant target dataset, unsupervised domain adaptation (UDA) has received considerable interests in recent years. Existing methods can be broadly classified into three categories: discrepancy based, reconstruction based, and adversarial based. Discrepancy based methods usually introduce a divergence criterion to measure the distance between the source and target data distributions, and then achieve model adaptation by minimizing the corresponding criterion, e.g. the maximum mean discrepancy (MMD)~\cite{MMD}, the wasserstein metric~\cite{wasserstein}, and the contrastive domain discrepancy~\cite{constrast_da}. Reconstruction based methods~\cite{recon_da_1,recon_da_2,recon_da_3} typically introduce an auxiliary image reconstruction task that guides the network to extract domain-invariant features for model adaptation. Inspired by GAN, there are also approaches~\cite{dann,cdan,dirt} that introduce domain discriminators to learn domain-invariant features in an adversarial manner. Despite of effectiveness, these methods require access to the source data, which is often impractical due to data privacy or security concerns.

\subsection{Source-free Domain Adaptation}
\par Currently, there have been several works~\cite{3c-gan,gan_sfda,gan_sfss,shot,shot++,gsfda,nrc,ms_sfda} attempting to realize source-free domain adaptation, where only a pre-trained source model and unlabeled target data are available. In these approaches, \cite{3c-gan,gan_sfda,gan_sfss} introduce generative networks to generate pseudo-data similar to sources or targets, which are difficult and inefficient. Instead, self-training based on feature prototype could be a promising direction~\cite{shot,shot++,pct,ms_sfda}. The most relevant papers to our BMD are SHOT~\cite{shot} and SHOT++~\cite{shot++}, which introduce a weighted k-means clustering algorithm to generate feature prototype and then assign pseudo-labels based on prototype matching. However, their feature prototype generation process is category-biased and prone to introducing negative transfer for those hard-transfer data. In contrast, our BMD can obtain more robust and stable feature prototypes by introducing inter-class balanced sampling and intra-class multicentric prototypes generation. {Our strategy may share some similarities in other fields approaches{~\cite{noisy_learning,PCN}}. However, our methods are fundamentally different. Unlike these works which apply the labeled data to train a network to recognize the multimodal classes by introducing the multicentric prototypes, we explore the multicentric idea to assign preciser pseudo labels for those unlabeled data, especially those hard-transfer data, to achieve source-free model adaptation.}

\subsection{Imbalanced Learning}

\par Massive studies~\cite{long_tail_survey} have been proposed for long-tail vision tasks to maintain diversity and balance predictions for minority categories. One straightforward idea is to perform category-balanced sampling~\cite{balance_learning_1,decouple,balance_sampling_1} using prior knowledge about the category distributions. However, these strategies are not applicable to the UDA task because we do not have access to knowledge of the target distribution. For DA methods, \cite{bnm,fast_bnm} introduce a nuclear-norm regularization item to achieve balanced learning, and \cite{cbst,crst} design a self-training framework to realize class-balance segmentation adaptation. In the absence of labeled data, DeepCluster~\cite{deepcluster}, one of the best self-supervised and class-balanced learning methods, generates pseudo-labels via k-means clustering and utilizes them to re-train the current model. Considering domain shift and absence of source data and leveraging advantages of existing methods, we design a novel and general class-balanced multicentric prototype strategy for SFDA.

\section{Preliminary}

\par In this paper, we consider the $K$-way object (2D images and 3D point cloud) recognition task. In the conventional UDA task, we are given two domain data, the labeled source domain with $n_s$ samples as $\mathcal{D}_s = \{(x^i_s, y^i_s)\}^{n_s}_{i=1}$ where $x^i_s \in \mathcal{X}_s$, $y_s^i \in \mathcal{Y}_s$ and $y_s^i$ is the corresponding label of data sample $x_s^i$, and the unlabeled target domain with $n_t$ samples as $\mathcal{D}_t = \{(x_t^i)\}^{n_t}_{i=1}$ where $x_t^i \in \mathcal{X}_
t$. The goal of UDA is to predict the labels $\{y_t^i\}^{n_t}_{i=1}$ of $\mathcal{D}_t$ where $y_t^i \in \mathcal{Y}_t$ with the well-labeled source domain $\mathcal{D}_s$. It is commonly assumed that the data space of $\mathcal{D}_s$ and $\mathcal{D}_t$ are distinct but label space are identical, i.e., $\mathcal{X}_s \neq \mathcal{X}_t$, $\mathcal{Y}_s = \mathcal{Y}_t$. But under the SFDA setting, the $\mathcal{D}_s$ is inaccessible and it is replaced by the source model. Assume that the source model $f_s$ has been well-trained and it consists of two parts: a feature extractor $g_s$: $\mathcal{X}_s \rightarrow \mathbb{R}^d$ and a classifier $h_s$: $\mathbb{R}^d \rightarrow \mathbb{R}^K$, i.e., $f_s(x) = h_s(g_s(x))$. Here $d$ is the dimension of the extracted feature. Therefore, the goal for SFDA is to learn the target model $f_t$: $\mathcal{X}_t \rightarrow \mathcal{Y}_t$ with only access to the source model $f_s$ and the unlabeled target domain $\mathcal{D}_t$.

\par To transfer the knowledge from the pre-trained source model, feature prototype based pseudo-labeling strategy is a promising direction. Inspired by DeepCluster~\cite{deepcluster}, existing feature prototype based pseudo-labeling strategies~\cite{shot,shot++,ms_sfda} first attain the prototype $c_k$ for each class similar to weighted k-means clustering as follows:
\begin{equation}
\begin{aligned}
    c_{k} &= \frac{\sum_{x_t \in \mathcal{X}_t} \delta_k(\hat{f}_t(x_t)) \hat{g}_t(x_t)}{\sum_{x_t \in \mathcal{X}_t}\delta_k(\hat{f}_t(x_t))},
\end{aligned}
\end{equation}
where $\hat{f}_t = \hat{g}_t \circ \hat{h}_t$ denotes the previously learned target model and $\delta_k(\hat{f}_t(x_t))$ denotes the softmax probability of target instance $x_t$ belonging to the $k$-th class. Then one can obtain the pseudo label $\hat{y}_t$ of target $x_t$ via the nearest prototype classifier as:
\begin{equation}
    \hat{y}_t = \arg \min_k D_f(\hat{g}_t(x_t), c_k),
\end{equation}
where $D_f(a, b)$ measures the distance between $a$ and $b$. One may iterate above process to obtain more stable prototype and pseudo labels like:
\begin{equation}
\begin{aligned}
	c_k &= \frac{\sum_{x_t\in \mathcal{X}_t}{\mathds{1}(\hat{y}_t=k)}\ \hat{g}_t(x_t)}{\sum_{x_t\in \mathcal{X}_t}{\mathds{1}(\hat{y}_t=k)}},\\
	\hat{y}_t &= \arg\min_k D_f(\hat{g}_t(x_t), c_k).
\end{aligned}
\end{equation}
where $\mathds{1}(\cdot)$ is an indicator function.Thereafter, based on the obtained pseudo labels, one can realize self-training based model adaptation with the categorical cross-entropy (CE) loss as follows:
\begin{equation}
    \mathcal{L}_{st} = - \frac{1}{n_t}\sum_{i=1}^{n_t}{\sum_{k=1}^{K} {\mathds{1}_{[k=\hat{y}_t]}}}\log \delta_k(f_t(x_t^i))).
\end{equation}


\section{BMD Strategy}
\label{sec:method}

\begin{figure}[h]
    \centering
  \includegraphics[width=0.60\linewidth]{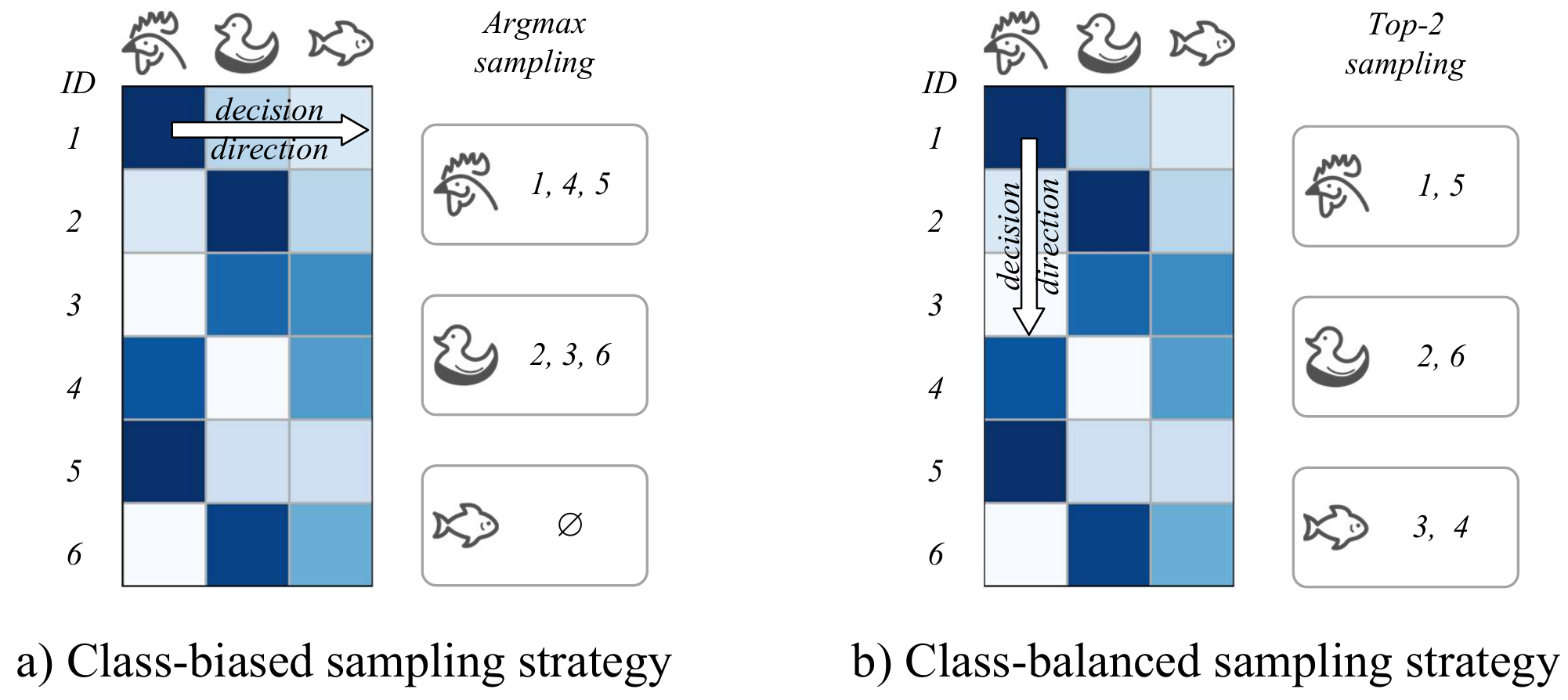}
  \caption{A toy example compares the existing class-biased strategy (left) with BMD class-balanced strategy (right). To better understand the difference, we illustrate the decision direction of these two sampling strategies. In the presence of large domain gaps, existing strategy is prone to aggregating class-biased data instances.}
    \label{fig:class_balanced_sampling}
\end{figure}

\subsection{Inter-class Balanced Prototype}

\par The visual domain gaps between source and target are typically different between categories, resulting in relatively higher prediction confidence scores for those easy-transfer classes in target domain~\cite{cbst,crst}. Therefore, we conjecture that existing strategies are category-biased and tend to generating noisy labels for those hard data. To avoid the gradual dominance of easy-transfer classes on prototype generation, we propose a novel global inter-class balanced sampling strategy to aggregate those potential data samples. Different from existing methods that decide whether to sample instances based on the instance-level prediction results, we formulate this as a multiple instance learning (MIL) problem~\cite{MIL,MIL_softbag}. In MIL, individual samples are grouped in two bags, i.e., positive and negative bags. A positive bag contains at least one positive instance and a negative bag contains no positive instance. For a specific class $k$, we treat the target domain $\mathcal{D}_t$ as a combination of a positive bag and a negative bag, where each data instance $x_t$ is represented by a feature vector $\hat{g}_t(x_t)$ and a classification result $p(x_t) = \delta(\hat{f}_t(x_t))$. Thus, the feature prototype for the $k$-th class is a representative of the positive bag. Since the \emph{top} instances are most likely to be positive, we then aggregate the top-$M$ $\delta_k(\hat{f}_t(x_t))$ scores represented instances along all target domain $\mathcal{D}_t$ for the $k$-th class as potential instances. After that we can average them to build the class-balanced feature prototype $c_{k}$ and assign the pseudo label $\hat{y}_t$ as:
\begin{equation}
\begin{aligned}
    \mathcal{M}_k &= \mathop{\arg \max}_{ x_t \in \mathcal{X}_t \atop |\mathcal{M}_k| = M} \delta_k(\hat{f}_t(x_t)),\\
    c_k &= \frac{1}{M} \sum_{i \in \mathcal{M}_k}{ \hat{g}_t(x_t^i)},\\
    \hat{y}_t &= \mathop{\arg \min}_k D_f(\hat{g}_t(x_t), c_k).
\end{aligned}
\end{equation}
where $M = \max\{1, \lfloor\frac{n_t}{r\times K}\rfloor \}$, $r$ is a hyperparameter denoting the top-$M$ selection ratio, and $K$ is the number of object classes in the target domain. For simplicity, we refer to this class-balanced sampling based feature prototype pseudo-labeling strategy as BP. It is worth noting that our BP strategy is not based on local instance-level prediction results to decide whether to sample instances, but rather to select the top-$M$ most likely instances to construct feature prototypes from a global perspective. Therefore, we argue that our strategy is inter-class balanced. We compare our class-balanced strategy with existing class-biased strategy in Fig.~\ref{fig:class_balanced_sampling} with a toy example. We may iterate this process like existing methods to obtain more stable prototypes and pseudo labels as:
\begin{equation}
\begin{aligned}
    \mathcal{M}_k &= \mathop{\arg \max}_{ x_t \in \mathcal{X}_t \atop |\mathcal{M}_k| = M} \frac{\exp{(\hat{g}_t(x_t) \cdot c_k)}}{\sum_{j=1}^{K} \exp{(\hat{g}_t(x_t)\cdot c_j)}},\\
    c_k &= \frac{1}{M} \sum_{i \in \mathcal{M}_k} \hat{g}_t(x_t^i),\\
    \hat{y}_t &= \mathop{\arg \min}_k D_f(\hat{g}_t(x_t), c_k).
\end{aligned}
\end{equation}

\subsection{Intra-class Multicentric Prototype}
\begin{figure}[b]
    \centering
    \includegraphics[width=0.65\textwidth]{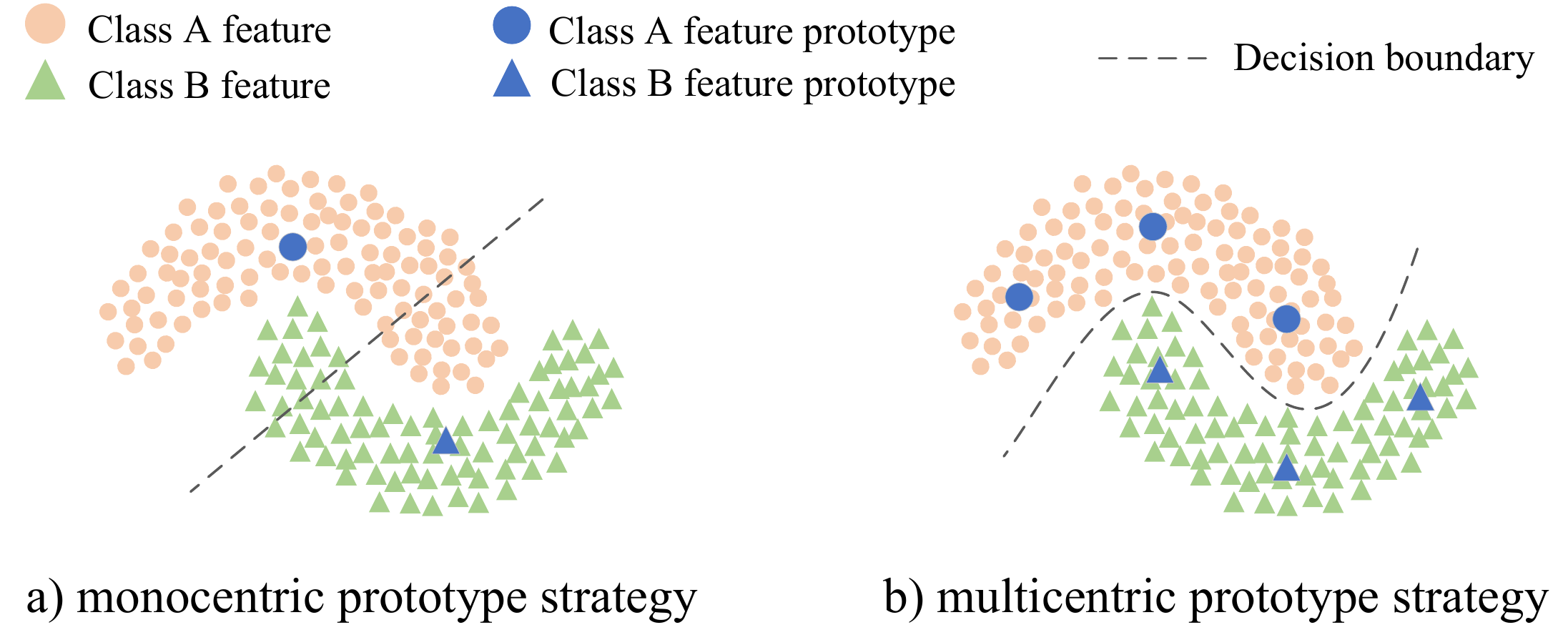}
    \caption{Comparison between existing monocentric prototype strategy (left) and our proposed multicentric prototype strategy (right). We can conclude that with multicentric prototype strategy we would obtain more robust and precise decision boundaries during pseudo labels generation.}
    \label{fig:multi_centric_strategy}
\end{figure}

\par Even though with the above strategy we can obtain class-balanced feature prototype and assign more robust pseudo labels, we found that a coarse monocentric feature prototype may be not effectively represent those ambiguous data and even introduce negative transfer. In contrast to \cite{proda} that introduces uncertainty for pseudo labels to mitigate the negative transfer caused by monocentric prototype, in this paper, we aim to assign more robust and precise label for each instance. Therefore, we propose an intra-class multicentric prototype strategy for each category to obtain more robust and precise pseudo label. We compare our multicentric strategy with existing monocentric strategy in Fig.~\ref{fig:multi_centric_strategy}.

\par Clustering is an essential data analysis technique for grouping unlabeled data in unsupervised learning~\cite{cluster_review}. In our implementations, assume the sampled data instances for $k$-th class as $\mathcal{X}_t^k$, the predefined multiple feature prototype number is $S$. We represent each data instance $x_t$ as the extracted $\hat{g}_t(x_t)$, and apply the classical $k$-means~\cite{kmeans} algorithm to realize intra-class clustering. Then for simplicity, we directly denote the $S$ cluster centroids $\{c_k^i\}_{i=1}^S$ as intra-class multiple feature prototypes for the $k$-th class. After obtaining multiple feature prototypes for all categories, we can assign the pseudo-labels as follows:
\begin{equation}
    \hat{y}_t = \mathop{\arg \max}_k \frac{\max\limits_{1\le i\le S}(\exp({\hat{g}_t(x_t) \cdot c_k^i}))}{\sum_{j=1}^K{\max\limits_{1\le i\le S}(\exp({\hat{g}_t(x_t) \cdot c_j^i}))}},
\end{equation}
where $c_k^i$ is the $i$-th feature prototype of class $k$. For simplicity, we refer to this class-balanced sampling based multicentric pseudo-labeling strategy as BMP. We may also iterate this process like before to obtain more stable prototype and pseudo labels as:
\begin{equation}
\begin{aligned}
    & \mathcal{M}_k = \mathop{\arg\max}_{x_t \in \mathcal{X}_t \atop|\mathcal{M}_k| = M} \frac{\max\limits_{1\le i\le S}(\exp({\hat{g}_t(x_t) \cdot  c_k^i}))}{\sum_{j=1}^K{\max\limits_{1\le i\le S}(\exp({\hat{g}_t(x_t) \cdot c_j^i}))}},\\
    & \{c_k^i\}_{i=1}^S = \mathop{Kmeans}_{n \in \mathcal{M}_k} (\hat{g}_t(x_t^n)),\\
    & \hat{y}_t = \mathop{\arg\max}_{k} \frac{\max\limits_{1\le i\le S}(\exp({\hat{g}_t(x_t) \cdot c_k^i}))}{\sum_{j=1}^K{\max\limits_{1\le i\le S}(\exp({\hat{g}_t(x_t) \cdot c_j^i}))}}.
\end{aligned}
\end{equation}

\subsection{Dynamic Pseudo Label}
\par Existing strategies update the pseudo label bank at a fixed training period, which may not effectively exploit the updated network during optimization. Thus, we further explore a dynamic pseudo-labeling strategy to improve the target model performance. At the beginning of each epoch, we first update the multiple feature prototype for each class and the corresponding pseudo labels for each instance from a global perspective. And then, for each iteration step we update the feature prototypes as the exponentially moving average (EMA)~\cite{EMA} of the cluster centroids in mini-batches. Specifically, we obtain the dynamic pseudo labels $\hat{y}_t^d$ and update the feature prototypes as follows:
\begin{equation}
    \begin{aligned}
        & \hat{y}_t^d = \frac{\max\limits_{1\le i\le S}(\exp({\hat{g}_t(x_t) \cdot c_k^i}))}{\sum_{j=1}^K{\max\limits_{1\le i\le S}(\exp({\hat{g}_t(x_t) \cdot c_j^i}))}},\\
        & p_k^i(x_t^n) = \frac{\exp(\hat{g}_t(x_t^n) \cdot c_k^i)}{\sum_{j=1}^{K}\sum_{s=1}^{S}\exp(\hat{g}_t(x_t^n) \cdot c_j^s)},\\
        & \hat{c}_k^i = \frac{\sum_{n=1}^N \hat{g}_t(x_t^n) \cdot  p_k^i(x_t^n)}{\sum_{n=1}^N  p_k^i(x_t^n)},\\
        & c^i_k \leftarrow \lambda c^i_k + (1 - \lambda) \hat{c}^i_k.
    \end{aligned}
\end{equation}
where $p_k^i(x_t)$ denotes the similarity of instance $x_t$ with existing feature prototypes, $\hat{c}_k^i$ represents the $i$-th feature prototype of class $k$ calculated with current training minibatch, and $\lambda$ is the momentum coefficient of EMA which we set to 0.9999. With the obtained dynamic pseudo labels $\hat{y}_t^d$, instead of using a standard cross-entropy loss, following~\cite{proda}, we adopt a more robust variant, symmetric cross-entropy loss (SCE)~\cite{SCE} to further enhance the noise-tolerance. Formally, the dynamic pseudo label loss is defined as follows:
\begin{equation}
\begin{aligned}
\mathcal{L}_{dym} = - &\frac{1}{N}\sum_{i=1}^{N}{\sum_{k=1}^{K}} \hat{y}_{t, k}^d \log \delta_k(f_t(x_t^i))) - \\
    &\frac{1}{N}\sum_{i=1}^{N}{\sum_{k=1}^{K}} \delta_k(f_t(x_t^i))) \log \hat{y}_{t, k}^d ,
\end{aligned}
\end{equation}
\par However, the dynamic pseudo label did not take into account the potential for a large domain shift and may lead to less informative class prototypes, especially when many samples are misclassified, as it is updated based on the features of the local minibatch. Therefore, we combine the static pseudo label based self-training loss with the dynamic loss to achieve more stable results as :
\begin{equation}
    \mathcal{L}_{bmd} = \alpha \mathcal{L}_{st} + \beta \mathcal{L}_{dym}.
\end{equation}
where $\alpha$ and $\beta$ are hyper-parameters to balance the two losses. Overall, we denote the combination of the class-balanced multicentric pseudo-labeling and dynamic pseudo labeling strategies as BMD.
\section{Experiment}
\label{sec:experiment}
\subsection{Experimental Setup}
\par We conduct extensive experiments to evaluate the effectiveness of our BMD strategy covering several popular benchmarks and representative methods below.
\subsubsection{Datasets}
\par We evaluate our BMD strategy on three 2D image and one 3D point cloud recognition benchmarks. \textbf{Office-31}~\cite{office31} is a standard benchmark that contains three domains (Amazon (\textbf{A}), DSLR (\textbf{D}), and Webcam (\textbf{W})) and each domains contains 31 object classes under the office environment.
\textbf{Office-Home}~\cite{officehome} is a challenging medium-sized benchmark that contains 4 domains (Real (\textbf{Rw}), Clipart (\textbf{Cl}), Art (\textbf{Ar}) and Product (\textbf{Pr})) with 65 classes and a total of 15,500 images. \textbf{VisDA-C}~\cite{visda} is a more challenging large-scale benchmark, which focus on 12-class synthetic-to-real object recognition tasks. Its source domain contains about 152k synthetic 3D object images while the target domain consists of 55k real object images sampled from Microsoft CoCo~\cite{coco}. \textbf{PointDA-10}~\cite{pointda} is the first 3D dataset designed for domain adaptation on point cloud, which contains three domains (ModelNet-10, ShapeNet-10 and ScanNet-10). There are about 27.7k training and 5.1k testing frame point clouds.
\subsubsection{Baselines}
\par We inject our BMD strategy to four existing SFDA methods to verify its versatility. \textbf{SHOT}~\cite{shot} proposes to freeze the source classifier and fine-tunes the source features extraction module by maximizing the mutual information and feature prototype based pseudo labels. \textbf{SHOT++}~\cite{shot++} extends the SHOT~\cite{shot} by introducing self-supervised learning for fine-tuning the feature extraction module and employing semi-supervised learning strategy to further improve the target domain performance. Different from SHOT and SHOT++ that introduce pseudo labeling strategy to realize model adaptation, \textbf{G-SFDA}~\cite{gsfda} and \textbf{NRC}~\cite{nrc} explore the local neighborhood structure of the target data in feature space to realize model adaptation. Although these two approaches do not introduce pseudo labeling, we find that our BMD strategy still fits seamlessly with these methods and consistently improves their performance.

\begin{table}[tbp]
  \centering
  \caption{Classification accuracies (\%) on small-sized Office-31 dataset with ResNet-50 as backbone. SF denotes source-free.}
  \addtolength{\tabcolsep}{2.0pt}
  \resizebox{0.95\textwidth}{!}{
    \begin{tabular}{lcccccccca}
    \toprule
    Method &Venue & SF    & A$\rightarrow$D  & A$\rightarrow$W  & D$\rightarrow$A  & D$\rightarrow$W  & W$\rightarrow$A  & W$\rightarrow$D  & \textbf{Avg} \\
    \midrule
    DANN~\cite{dann}& JMLR 2016 & \xmark & 79.7  & 82.0  & 68.2  & 96.9  & 67.4  & 99.1  & 82.2  \\
    \midrule
    CDAN~\cite{cdan} &NeurIPS 2018 & \xmark & 92.9  & 94.1  & 71.0  & 98.6  & 69.3  & 100.0  & 87.7  \\
    \midrule
    MDD~\cite{mdd} &ICML 2019 & \xmark & 93.5  & 94.5  & 74.6  & 98.4  & 72.2  & 100.0  & 88.9  \\
    \midrule
    GVB-GD~\cite{gvbgd} &CVPR 2020 & \xmark & 95.0  & 94.8  & 73.4  & 98.7  & 73.7  & 100.0  & 89.3  \\
    \midrule
    \midrule
    SHOT~\cite{shot} &ICML 2020 & \multirow{2}[2]{*}{\cmark} & 94.0  & 90.1  & 74.7  & 98.4  & 74.3  & 99.9  & 88.6  \\
    \textbf{SHOT  w/ BMD} &ours &      & 95.6  & 93.0  & 75.6  & 97.5  & 75.0  & 99.8  & \textbf{89.4} \\
    \midrule
    SHOT++~\cite{shot++}&TPAMI 2021 & \multirow{2}[2]{*}{\cmark} & 94.3  & 90.4  & 76.2  & 98.7  & 75.8  & 99.9  & 89.2  \\
    \textbf{SHOT++ w/ BMD}  &ours  &  & 96.2  & 94.2  & 76.0  & 98.0  & 76.0  & 100.0  & \textbf{90.1} \\
    \bottomrule
    \end{tabular}%
    }
  \label{tab:office31_compare}%
\end{table}%

\begin{table*}[tbp]
  \centering
  \caption{Accuracies (\%) on medium-sized Office-Home dataset with ResNet-50 as backbone.[* using our reproduced performance]}
  \addtolength{\tabcolsep}{-0.5pt}
    \resizebox{0.97\textwidth}{!}{
    \begin{tabular}{lcccccccccccccca}
    \toprule
    Methods & Venue & SF & Ar$\rightarrow$ Cl & Ar$\rightarrow$Pr & Ar$\rightarrow$Re & Cl$\rightarrow$Ar & Cl$\rightarrow$Pr & Cl$\rightarrow$Re & Pr$\rightarrow$Ar & Pr$\rightarrow$Cl & Pr$\rightarrow$Re & Re$\rightarrow$Ar & Re$\rightarrow$Cl & Re$\rightarrow$Pr & \textbf{Avg} \\
    \midrule
    CDAN~\cite{cdan}  & NeurIPS 2018 & \xmark & 50.7  & 70.6  & 76.0  & 57.6  & 70.0  & 70.0  & 57.4  & 50.9  & 77.3  & 70.9  & 56.7  & 81.6  & 65.8  \\
    \midrule
    CDAN+BNM~\cite{bnm} & CVPR 2020 & \xmark & 56.2  & 73.7  & 79.0  & 63.1  & 73.6  & 74.0  & 62.4  & 54.8  & 80.7  & 72.4  & 58.9  & 83.5  & 69.4  \\
    \midrule
    GVB-GD~\cite{gvbgd} & CVPR 2020 & \xmark & 57.0  & 74.7  & 79.8  & 64.6  & 74.1  & 74.6  & 65.2  & 55.1  & 81.0  & 74.6  & 59.7  & 84.3  & 70.4  \\
    \midrule
    Fixbi~\cite{fixbi} & CVPR 2021 & \xmark & 58.1  & 77.3  & 80.4  & 67.7  & 79.5  & 78.1  & 65.8  & 57.9  & 81.7  & 76.4  & 62.9  & 86.7  & 72.7  \\
    \midrule
    \midrule
    SHOT*~\cite{shot} & ICML 2020 & \multirow{2}[2]{*}{\cmark} & 54.3  & 78.1  & 80.3  & 68.3  & 79.1  & 80.1  & 68.7  & 54.1  & 82.0  & 73.1  & 57.0  & 83.0  & 71.5  \\
    \textbf{SHOT w/ BMD} & ours  &       & 55.9  & 77.8  & 80.8  & 69.7  & 79.3  & 79.9  & 69.6  & 56.6  & 82.6  & 73.3  & 59.5  & 85.1  & \textbf{72.5} \\
    \midrule
    G-SFDA*~\cite{gsfda} & ICCV 2021 & \multirow{2}[1]{*}{\cmark} & 55.2  & 77.6  & 80.1  & 67.7  & 75.6  & 79.1  & 66.3  & 54.8  & 81.6  & 72.5  & 58.1  & 84.0  & 71.0  \\
    \textbf{G-SFDA w/ BMD} & ours  &       & 56.0  & 78.2  & 80.4  & 69.1  & 79.0  & 79.4  & 67.5  & 55.8  & 82.4  & 73.7  & 58.7  & 83.8  & \textbf{72.0} \\
    \midrule
    SHOT++*~\cite{shot++} & TPAMI 2021 & \multirow{2}[1]{*}{\cmark} & 55.9  & 79.1  & 81.8  & {69.9}  & {81.3}  & {81.0}  & 70.3  & 56.2  & 83.6  & 72.9  & 59.0  & 84.3  & {72.9} \\
    \textbf{SHOT++ w/ BMD} & ours  &       &{58.1}  & {79.7}  & {82.6}  & 69.3  & 81.0  & 80.7  & {70.8}  & {57.6}  & {83.6}  & {74.0}  & {60.0}  & {85.9}  & \textbf{73.6} \\
    \bottomrule
    \end{tabular}%
    }
  \label{tab:officehome_compare}%
\end{table*}%

\subsubsection{Implementation Details}
\par For a fair comparison, we adopt the same network architecture and training recipe with baselines. Specifically, we adopt the ResNet-50~\cite{resnet} pretrained on ImageNet~\cite{imagenet} as backbone for Office-31 and Office-Home benchmarks, and the ResNet-101 for VisDA-C benchmark. As for PointDA-10, we utilize the PointNet~\cite{pointnet} with local node aggregation network proposed in~\cite{pointda} as feature extraction backbone. To prepare the pretrained source model, following SHOT and NRC, we utilize the label smoothing~\cite{labelsmooth} to increase the discriminability of the source model and facilitate the following target data alignment. During target model adaptation, to achieve source and target domain alignment, we fix the target classifier $h_t = h_s$ and update only the target feature extractor $g_t$ initialized from $g_s$. Following previous methods, we apply the SGD optimizer with momentum 0.9 and the Adam optimizer for PointDA-10. The batch size is set to 64 for all benchmark datasets. We set the learning rate to 1e-2 for Office-31 and OfficeHome, 1e-3 for VisDA-C, and 1e-6 for PointDA-10. We train 30 epochs for all 2D image datasets and 50 epochs for PointDA. We set the hyperparamter $r$ to 3 for all datasets, and $S=4$ for Office-Home and VisDA-C, $S=2$ for PointDA-10 and Office-31. We set $\alpha=2$ and $\beta=0.5$ for VisDA-C, $\alpha=0.3$ and $\beta=0.1$ for Office-31 and Office-Home, and $\alpha=1.0$ and $\beta=0.1$ PointDA-10. All experiments are conducted on a RTX-3090 GPU with PyTorch-1.7.

\subsection{Results}
\subsubsection{2D Image Recognition}
\par We first evaluate the effectiveness of our strategy with existing methods on three 2D image recognition datasets. The results are summarized in Table~\ref{tab:office31_compare}-\ref{tab:visda_compare}, the top part illustrates results for the traditional UDA methods with access to source data during model adaptation, and the bottom part presents results for the SFDA methods. As shown in Table~\ref{tab:office31_compare}, on \textbf{Office-31}, our BMD strategy can consistently improve SHOT and SHOT++ to yield new state-of-the-art performance, especially on the challenging A $\rightarrow$ D task, our BMD strategy can improve SHOT from 90.1\% to 93.0\% and SHOT++ from 90.4\% to 94.2 \%, respectively. As excepted in Table~\ref{tab:officehome_compare}, on the medium-sized \textbf{Office-Home}, our BMD strategy can also consistently improve existing state-of-the-art methods. Specifically, by injecting BMD strategy, we can improve SHOT from 71.5\% to 72.5\%, G-SFDA from 71.0\% to 72.0\%, SHOT++ from 72.9\% to 73.6\%. For the large-scale synthetic-to-real \textbf{VisDA-C} dataset in Table~\ref{tab:visda_compare}, our BMD strategy can also significantly improve existing methods by a large margin, especially, we can improve SHOT from 82.9\% to 85.7\%, G-SFDA from 84.8\% to 86.5\%, NRC from 85.9\% to 86.9\% and SHOT++ from 87.3\% to 88.7\%. With our BMD strategy, SHOT++ can even achieve a performance comparable to the target supervised approach (88.7\% vs 89.6\%). In addition, on VisDA-C, we can find that with our strategy above methods can achieve more class-balanced performance. Especially for the challenging class ‘truck’, we can significantly improve SHOT from 58.2\% to 70.8\%, G-SFDA from 44.8\% to 59.7\%, and SHOT++ from 28.8\% to 45.9\%. We also report the standard deviation $\sigma$ of the accuracy achieved by our method. The $\sigma$ of SHOT w/BMD on Office-31 is 0.07, while the $\sigma$ of SHOT w/BMD on VisDA-C is 0.11, showing that the improvement of using BMD in SHOT is significant. Beyond closet-set SFDA, we further evaluate BMD with SHOT on two other DA scenarios, multi-source~\cite{multi_source} and multi-target~\cite{multi_target}. Due to space limitations, we will present these experiments in the Appendix.

\begin{table*}[htbp]
  \centering
  \caption{Per-class accuracy (\%) on large-scale VisDA-C validation set with ResNet-101 as backbone.}
  \addtolength{\tabcolsep}{-0.0pt}
  \resizebox{0.97\textwidth}{!}{
    \begin{tabular}{lcccccccccccccca}
    \toprule
    Methods & Venue & SF & plane & bcycl & bus   & car   & horse & knife & mcycl & person & plant  & sktbrd & train & truck & \textbf{Avg} \\
    \midrule
    CDAN~\cite{cdan}  & NeurIPS 2018 & \xmark & 85.2  & 66.9  & 83.0  & 50.8  & 84.2  & 74.9  & 88.1  & 74.5  & 83.4  & 76.0  & 81.9  & 38.0  & 73.9  \\
    \midrule
    SWD~\cite{swd}   & CVPR 2019 & \xmark & 90.8  & 82.5  & 81.7  & 70.5  & 91.7  & 69.5  & 86.3  & 77.5  & 87.4  & 63.6  & 85.6  & 29.2  & 76.4  \\
    \midrule
    MCC~\cite{mcc}   & ECCV 2020 & \xmark & 88.7  & 80.3  & 80.5  & 71.5  & 90.1  & 93.2  & 85.0  & 71.6  & 89.4  & 73.8  & 85.0  & 36.9  & 78.8  \\
    \midrule
    STAR~\cite{star}  & CVPR 2020 & \xmark &95.0 &84.0 &84.6 &73.0 &91.6 &91.8 &85.9 &78.4 &94.4 &84.7 &87.0 &42.2 & 82.7\\
    \midrule
    FixBi~\cite{fixbi}   & CVPR 2021 & \xmark &96.1 &87.8 &90.5 &90.3 &96.8 &95.3 &92.8 &88.7 &97.2 &94.2 &90.9 &25.7 &87.2  \\
    \midrule
    \midrule
    SHOT~\cite{shot}  & ICML 2020 & \multirow{2}[2]{*}{\cmark} & 94.3  & 88.5  & 80.1  & 57.3  & 93.1  & 94.9  & 80.7  & 80.3  & 91.5  & 89.1  & 86.3  & 58.2  & 82.9  \\
    \textbf{SHOT w/ BMD} & ours  &       & 96.2  & 87.8  & 81.4  & 61.7  & 95.0  & 97.5  & 87.9  & 82.9  & 92.6  & 88.8  & 87.4  & {70.8}  & \textbf{85.8} \\
    \midrule
    G-SFDA~\cite{gsfda} & ICCV 2021 & \multirow{2}[2]{*}{\cmark} & 95.9  & 88.1  & 85.4  & 72.5  & 96.1  & 93.7  & 88.5  & 80.6  & 92.3  & 92.2  & 87.6  & 44.8  & 84.8 \\
    \textbf{G-SFDA w/ BMD}  & ours  &       & 95.9  & 87.5  & 83.9  & 75.7  & 96.5  & 96.6  & 91.4  & 81.8  & 95.9  & 88.4  & 85.1  & {59.7}  & \textbf{86.5} \\
    \midrule
    NRC~\cite{nrc}   & NeurIPS 2021 & \multirow{2}[2]{*}{\cmark} & 96.8  & 91.3  & 82.4  & 62.4  & 96.2  & 95.9  & 86.1  & 80.6  & 94.8  & 94.1  & 90.4  & 59.7  & 85.9  \\
    \textbf{NRC w/ BMD} & ours  &       & 96.7  & 87.2  & 85.0  & 75.6  & 96.8  & 97.0  & 91.6  & 84.9  & 94.7  & 89.0  & 88.6  & 55.6  & \textbf{86.9} \\
    \midrule
    SHOT++~\cite{shot++} & TPAMI 2021 & \multirow{2}[2]{*}{\cmark} & {97.7}  & {88.4}  & {90.2}  & 86.3  & {97.9}  & {98.6}  & {92.9}  & 84.1  & {97.1}  & 92.2  & {93.6}  & 28.8  & 87.3  \\
    \textbf{SHOT++ w/ BMD} & ours  &       & 96.9  & 87.8  & 90.1  & {91.3}  & 97.8  & 97.8  & 90.6  & {84.4}  & 96.9  & {94.3}  & 90.9  & {45.9}  & \textbf{88.7} \\
    \midrule
    \midrule
    Target-Supervised & \textbf{\_} & \textbf{\_} & 97.0  & 86.6  & 84.3  & 88.7  & 96.3  & 94.4  & 92.0  & 89.4  & 95.5  & 91.8  & 90.7  & 68.7  & {89.6} \\
    \bottomrule
    \end{tabular}%
    }
  \label{tab:visda_compare}%
\end{table*}%

\subsubsection{3D Point Cloud Recognition}
\par In addition to 2D images, to verify the generality of our BMD strategy, we also conducted experiments on the 3D point cloud \textbf{PointDA-10} dataset. As shown in Table~\ref{tab:pointda_compare}, our BMD strategy can also significantly improve existing methods by a large margin. Specifically, by injecting BMD strategy, we can improve SHOT from 53.1\% to 57.0\% and NRC from 52.6\% to 57.0\%. Especially on the challenging task, SH $\rightarrow$ M, we can improve SHOT from 75.8\% to 81.5\% and NRC from 59.8\% to 83.4\%.

\begin{table}[t]
  \centering
  \addtolength{\tabcolsep}{2.0pt}
  \caption{Accuracies (\%) on PointDA-10 dataset with PointNet~\cite{pointnet} as backbone.}
  \resizebox{0.90\textwidth}{!}{
    \begin{tabular}{lcccccccca}
    \toprule
    Method &Venue & SF   & M$\rightarrow$SC & M$\rightarrow$SH & SC$\rightarrow$M & SC$\rightarrow$SH & SH$\rightarrow$M & SH$\rightarrow$SC & \textbf{Avg}. \\
    \midrule
    ADDA~\cite{ADDA} &CVPR 2017 &\xmark & 30.5  & 61.0  & 48.9  & 51.1  & 40.4  & 29.3  & 43.5  \\
    \midrule
    MCD~\cite{MCD} &CVPR 2018 &\xmark & 31.0  & 62.0  & 46.8  & 59.3  & 41.4  & 31.3  & 45.3  \\
    \midrule
    PointDAN~\cite{pointda} &NeurIPS 2019 &\xmark & 33.0  & 64.2  & 49.1  & 64.1  & 47.6  & 33.9  & 48.7  \\
    \midrule
    \midrule
    VDM~\cite{VDM}  &Arxiv 2021 &\cmark & 30.9  & 58.4  & 45.3  & 61.8  & 61.0  & 40.8  & 49.7  \\
    \midrule
    SHOT~\cite{shot} &ICML 2020 &\multirow{2}[2]{*}{\cmark} & 31.8  & 62.1  & 67.6  & 56.9  & 75.8  & 24.3  & 53.1  \\
    \textbf{SHOT w/ BMD} &ours &      & 32.8  & 66.1  & 75.0  & 62.0  & 81.5  & 24.4  & \textbf{57.0} \\
    \midrule
    NRC~\cite{nrc} &NeurIPS 2021 &\multirow{2}[2]{*}{\cmark} & 25.8  & 64.8  & 70.1  & 68.1  & 59.8  & 26.9  & 52.6  \\
    \textbf{NRC w/ BMD} &ours & & 33.8  & 66.7  & 70.8  & 62.6  & 83.4  & 24.8  & \textbf{57.0} \\
    \bottomrule
    \end{tabular}%
    }
  \label{tab:pointda_compare}%
\end{table}%

\subsection{Performance Analysis}
\subsubsection{Ablation Study}

\begin{table}[tp]
\begin{minipage}{.47\linewidth}
  \centering
  \caption{Ablation study on three UDA datasets.}
  \addtolength{\tabcolsep}{-0.0pt}
  \resizebox{1.0\textwidth}{!}{
    \begin{tabular}{lccc}
    \toprule
    Methods/Datasets &Office-Home & VisDA-C & PointDA-10\\
    \midrule
    Source-model  & 59.6  & 46.6 & 39.4\\
    SHOT w/ naive PL~\cite{psd_label}  & 70.3   &82.9  &51.0 \\
    SHOT w/ mono PL~\cite{shot}        & 71.5   &82.9  &53.1\\
    \textbf{SHOT w/ BP (ours)} & 72.0  &83.8  &55.0 \\
    \textbf{SHOT w/ BMP (ours)} & 72.5 &84.7 &56.4\\
    \textbf{SHOT w/ BMD (ours)} & 72.5 &85.7 &57.0\\ 
    \bottomrule
    \end{tabular}%
    }
  \label{tab:ablation}%
\end{minipage}
\ \ \ 
\begin{minipage}{.47\linewidth}
  \centering
  \caption{Statistics of class-wise performance on VisDA-C.}
  \resizebox{1.0\textwidth}{!}{
    \begin{tabular}{lccc}
    \toprule
    Methods & Acc avg $\mu$ $\uparrow$ & Acc std $\sigma$ $\downarrow$ & Acc cv $c_v  \downarrow$ \\
    \midrule
    SHOT~\cite{shot}  & 82.9  & 12.857 & 0.155  \\
    \textbf{SHOT w/ BMD} & 85.8  & 10.127 & \textbf{0.118}  \\
    \midrule
    G-SFDA~\cite{gsfda} & 84.8  & 14.279 & 0.168  \\
    \textbf{G-SFDA w/ BMD} & 86.5  & 10.766 & \textbf{0.124} \\
    \midrule
    SHOT++~\cite{shot++} & 87.3  & 19.027 & 0.218  \\
    \textbf{SHOT++ w/ BMD} & 88.7  & 14.146 & \textbf{0.159}  \\
    \bottomrule
    \end{tabular}%
    }
  \label{tab:variance}%
\end{minipage}
\end{table}%

\par As we presented before, the core components of our BMD strategy are inter-class balanced sampling and intra-class multicentric prototype based pseudo label generation. In order to incorporate the network dynamic optimization information, we further introduce the EMA based dynamic pseudo label strategy. To study the advantage of each part of our BMD strategy, we conduct the ablation study on Office-Home, VisDA-C and PointDA-10 with SHOT, the results are summarized in Table~\ref{tab:ablation}. To verify the superiority of our BMD strategy, we also introduce two existing strategies, the naive argmax based pseudo label~\cite{psd_label} strategy, and the monocentric prototype based pseudo label~\cite{shot} strategy. As expected, the results show that the simple BP strategy can outperform existing strategies, which indicates the importance of the class-balanced sampling strategy for pseudo label generation. When we incorporate the intra-class multicentric prototype strategy with BP strategy, i.e. the BMP, the performance is further significantly boosted. We attribute this to the fact that the MP strategy introduce more fine-grained feature prototypes for each class, which allows the model to assign more accurate pseudo-labels for those hard-transfer data. As for the dynamic strategy, we find that it is not as effective as BP and BMP on the Office-Home and PointDA datasets. We suspect it may be due to the relatively small size of the datasets, thus the EMA-based dynamic feature prototypes cannot effectively utilize the information during training. Due to space limitations, we presented more ablation experiments in the Appendix.

\subsubsection{Does our strategy really achieve more class-balanced results?}

\par To verify whether our BMD strategy is really helpful in achieving the class-balanced results, in this part, we introduce the coefficient of variation (also known as the relative standard deviation) $c_v$ as metric to evaluate the inter-class balance performance, which is a standardized measure of dispersion of a probability distribution or frequency distribution. Formally, the coefficient of variation is defined as $c_v = \frac{\sigma}{\mu}$, where $\sigma$ and $\mu$ are the standard deviation and expected mean of the data distribution, respectively. We conduct experiments on VisDA-C dataset, the results are summarized in Table~\ref{tab:variance}. As shown in this table, for all methods by injecting our BMD strategy we can arrive higher accuracy mean $\mu$, lower standard deviation $\sigma$ and lower coefficient of variation $c_v$, which demonstrates that our BMD strategy indeed facilitates existing methods to achieve more class-balanced performance. To verify the robustness, we also conducted the $c_{v}$ experiments on PointDA, on task SC$\rightarrow$M, SHOT gets 0.291, SHOT w/BMD is 0.186; NRC gets 0.464, NRC w/BMD is 0.301. These results further demonstrate that our BMD strategy can improve the existing methods to achieve class-balanced results.

\subsubsection{Visualization}
\begin{figure}[t]
    \centering
    \includegraphics[width=0.85\textwidth]{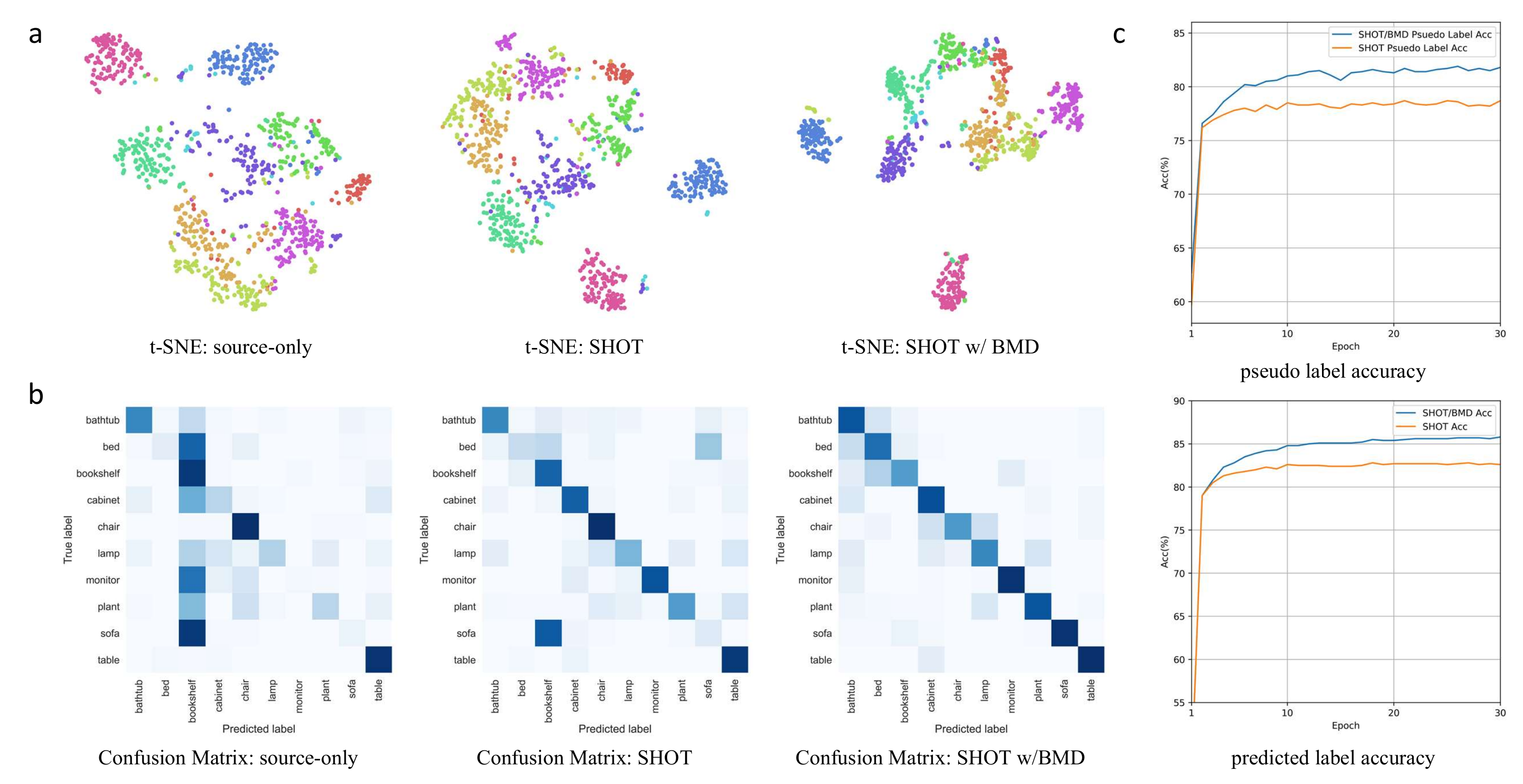}
    \caption{\textbf{a}: The t-SNE visualization of target features for source model, SHOT, and SHOT w/ BMD on PointDA-10 (SC$\rightarrow$M). \textbf{b}: The Confusion Matrix visualization for source model, SHOT, and SHOT w/BMD on PointDA-10 (SC$\rightarrow$M). \textbf{c}: The pseudo and predicted label accuracy curves for SHOT, and SHOT w/BMD on VisDA-C.}
    \label{fig:tsne}
\end{figure}

\par {To demonstrate the superiority of our BMD strategy, we present the t-SNE feature and confusion matrix on PointDA-10 (SC$\rightarrow$M), and the pseudo and predicted label accuracy curves on VisDA-C in Fig.~{\ref{fig:tsne}}. From Fig.~{\ref{fig:tsne}}a, we can see that after model adaptation by SHOT w/ BMD, the target features are more compactly clustered. The confusion matrix in Fig.~{\ref{fig:tsne}}b demonstrates that our BMD strategy can achieve more class-balanced accuracy compared to the vanilla SHOT. In particular, when there is severe class bias in the source model, e.g., for the hard-transfer \emph{`sofa'} class, SHOT cannot achieve good model adaptation due to the severe domain gap, while BMD strategy can overcome this well by our inter-class balanced sampling. The accuracy curves in Fig.~{\ref{fig:tsne}}c further support that our BMD strategy can facilitate existing methods to achieve more superior pseudo labels and predicted labels during model adaptation.}

\section{Conclusion}
\label{sec:conclusion}

\par In this paper, we present a general class-balanced multicentric dynamic (BMD) prototype strategy for source-free domain adaptation, which is model agnostic and can be applied to existing self-training based SFDA methods. Specifically, our BMD strategy consists of a novel inter-class balanced sampling strategy, an intra-class multicentric prototype strategy, and a dynamic feature prototype based pseudo-labeling strategy. We have injected our strategy into four existing representative methods and conducted experiments on both 2D images and 3D point cloud datasets. The results demonstrate that our BMD strategy can consistently and significantly boost existing methods to yield new state-of-the-art performance. For future work, we will extend our BMD strategy to those source-free dense prediction tasks. 

\par \noindent \textbf{Acknowledgments}: This work was supported by Shanghai Municipal Science and Technology Major Project (No.2018SHZDZX01), ZJ Lab, Shanghai Center for Brain Science and Brain-Inspired Technology, and the National Natural Science Foundation of China under Grant 61906138, by Shanghai Rising Star Program (No. 21QC1400900).
\bibliographystyle{splncs04}
\bibliography{egbib.bib}

\end{document}